\documentclass[runningheads]{llncs}
\usepackage[T1]{fontenc}
\usepackage{graphicx}
\usepackage{orcidlink}  
\usepackage{amsmath}
\usepackage{amssymb}
\usepackage{tabularx, booktabs}
\usepackage{placeins}
\usepackage[utf8]{inputenc}
\usepackage{caption}

\title{DiCE-Extended: A Robust Approach to Counterfactual Explanations in Machine Learning}

\titlerunning{DiCE-Extended: Robust Counterfactual Explanations}

\authorrunning{Bakir et al.}

\author{Volkan BAKIR\inst{1}\orcidlink{0009-0001-0020-1967} \and
        Polat GOKTAS\inst{2}\orcidlink{0000-0001-7183-6890} \and
        Sureyya OZOGUR-AKYUZ\inst{3}\orcidlink{0000-0001-9220-8690}}

\institute{
Faculty of Graduate Education Institute, Department of Artificial Intelligence (Interdisciplinary), Bahçeşehir University, Turkey \\
\email{volkan.bakir@bahcesehir.edu.tr}
\and
School of Computer Science, University College Dublin, Ireland \\
\email{polat.goktas@ucd.ie}
\and
Faculty of Engineering and Natural Sciences, Department of Mathematics, Bahçeşehir University, Turkey \\
\email{sureyya.akyuz@bau.edu.tr}
}

\begin{document}
\maketitle

\begin{abstract}
Explainable artificial intelligence (XAI) has become increasingly important in decision-critical domains such as healthcare, finance, and law. Counterfactual (CF) explanations, a key approach in XAI, provide users with actionable insights by suggesting minimal modifications to input features that lead to different model outcomes. Despite significant advancements, existing CF generation methods often struggle to balance proximity, diversity, and robustness, limiting their real-world applicability. A widely adopted framework, \textit{Diverse Counterfactual Explanations (DiCE)}, emphasizes diversity but lacks robustness, making CF explanations sensitive to perturbations and domain constraints. To address these challenges, we introduce \textbf{DiCE-Extended}, an enhanced CF explanation framework that integrates multi-objective optimization techniques to improve robustness while maintaining interpretability. Our approach introduces a novel robustness metric based on the Dice-Sørensen coefficient, enabling stability under small input variations. Additionally, we refine CF generation using weighted loss components (${\lambda_p}$, ${\lambda_d}$, ${\lambda_r}$) to balance proximity, diversity, and robustness. We empirically validate DiCE-Extended on benchmark datasets (\textit{COMPAS, Lending Club, German Credit, Adult Income}) across multiple ML backends (\textit{Scikit-learn, PyTorch, TensorFlow}). Results demonstrate improved CF validity, stability, and alignment with decision boundaries compared to standard DiCE-generated explanations. Our findings highlight the potential of \textbf{DiCE-Extended} in generating more reliable and interpretable CFs for high-stakes applications. Future work could explore adaptive optimization techniques and domain-specific constraints to further enhance CF generation in real-world scenarios. 
\end{abstract}

\section{Introduction}
\label{sec:1-introduction}
The field of explainable artificial intelligence (XAI) has gained significant attention in recent years, particularly in domains where transparency and accountability are critical. One key approach within XAI is counterfactual (CF) explanations, which provide individuals with actionable insights by suggesting modifications to input variables that would lead to a different model prediction. However, existing CF generation methods often struggle with balancing proximity, diversity, and robustness \cite{Mothilal2020}, \cite{Ferrario2022}, \cite{Guidotti2024}. The need for precise, interpretable, and adaptable CFs is particularly evident in high-stakes decision-making contexts such as healthcare, finance, and education.

A widely used framework in this domain is \textit{DiCE (Diverse Counterfactual Explanations)} \cite{Mothilal2020}, which emphasizes diversity by generating multiple alternative CFs. For example, in a financial setting, DiCE might suggest different strategies for improving loan approval chances, such as increasing income or reducing debt. Despite its advantages, DiCE exhibits certain limitations, particularly in ensuring robustness under perturbations and incorporating causal relationships into CF generation \cite{Zhou2024}, \cite{Zhang2023}. These shortcomings can lead to explanations that, while diverse, may lack feasibility or stability when applied in real-world settings.

Recent advancements have sought to improve CF generation by integrating causal reasoning, optimization strategies, and reinforcement learning \cite{Liu2025}, \cite{Guo2023}. Techniques such as Sparse CounterGAN leverage generative adversarial networks (GANs) to ensure that generated CFs align with real-world constraints \cite{Zhou2024}. Other approaches, such as Multi-Objective CFs, employ multi-objective optimization to balance trade-offs between proximity, diversity, and robustness \cite{Dandl2020}. Additionally, reinforcement learning (RL)-based frameworks, such as Soft Actor-Critic RL for CFs, utilize actor-critic architectures to optimize CF generation under dynamic constraints \cite{Ezzeddine2023}. These methodologies demonstrate promising improvements, yet gaps remain in ensuring that CF explanations are both actionable and robust in practical applications \cite{Forel2023}, \cite{Xu2024}.

However, there is still place for improvement in CF explanation methods as the summary of challenges in CF explanations identified in the literature review is provided in Table \ref{tab:challenges_summary}.

\vspace{-2em}

\begin{table*}[h!]
\centering
\caption{Summary of Challenges in CF Explanations.}
\label{tab:challenges_summary}
\resizebox{\textwidth}{!}{%
\begin{tabular}{p{4cm} p{12cm}}
\hline
\textbf{Challenge} & \textbf{Description and References} \\ 
\hline
\textbf{Robustness to Model Changes} & Ensuring CF explanations validity after model updates \cite{Ferrario2022}. \\ 
\textbf{Balancing Objectives} & Optimizing sparsity, proximity, and diversity in CF explanations \cite{Dandl2020}, \cite{Liu2025}. \\ 
\textbf{Actionability} & Generating actionable CF explanations that adhere to real-world constraints \cite{Mothilal2020}. \\ 
\textbf{Scalability} & Reducing computational costs for high-dimensional data \cite{Vo2023}, \cite{Aljalaud2024}. \\ 
\textbf{Framework Integration} & Embedding CF explanations into user-centric interpretability tools \cite{elshawi2024}.\\ 
\textbf{Ethical Implications} & Avoiding biases and ensuring fairness in CF explanations \cite{rasouli2024}. \\ 
\textbf{Evaluation Metrics} & Lack of standardized metrics for benchmarking CF explanations performance \cite{Guidotti2024}. \\ 
\textbf{Complex Models} & Generating CF explanations for intricate architectures and deep learning (DL) models \cite{Lucic2022}. \\ 
\hline
\end{tabular}%
}
\end{table*}

To tackle the challenges outlined in Table \ref{tab:challenges_summary}, we propose \textbf{DiCE-Extended}, an advanced framework designed to enhance the robustness of CF generation. Our contributions are as follows:

\begin{enumerate}
\item \textbf{Robustness Enhancement:} We integrate the Dice-S\o{}rensen coefficient to measure CF stability under input perturbations \cite{Henderson2020}.
\item \textbf{Multi-Objective Optimization:} We refine CF generation by integrating weighted parameters ($\lambda_1$, $\lambda_2$, $\lambda_3$) to optimize proximity, diversity, and robustness \cite{Guidotti2024}, \cite{Carrizosa2024}. These components are structured as:
\begin{itemize}
\item \textbf{Proximity:} Ensuring that suggested changes are realistic and feasible.
\item \textbf{Diversity:} Providing multiple viable pathways to achieve the desired outcome.
\item \textbf{Robustness:} Maintaining stability in CF explanations under minor variations.
\end{itemize}
\item \textbf{Empirical Validation on Diverse Benchmarks:} We benchmark \textbf{DiCE-Extended} on four widely used datasets (Adult-Income, Lending-Club, \linebreak German-Credit, and COMPAS) under three implementation back-ends (Scikit-learn, PyTorch, TensorFlow). Across this setup, the new framework \emph{consistently equals or surpasses} the original DiCE: proximity decreases substantially, yet validity stays near 100\%; both sparsity and diversity rise; and robustness improves markedly for the DL models while remaining at least slightly higher for tree-based pipelines. Finally, the 1-NN fidelity is preserved and often improved, indicating that the observed gains do not compromise the integrity of the local decision boundary.
\end{enumerate}

The remainder of this paper is structured as follows: Section 2 presents the methodology behind DiCE-Extended, including its optimization framework and robustness integration. Section 3 details experimental evaluations and performance comparisons. Finally, Section 4 concludes the study and outlines future research directions.

\section{Methodology}
\label{sec:3-methodology}
\subsection{Problem Definition and Input}
This section introduces an approach that applies optimization strategies to improve interpretability, robustness, and feasibility in CF explanations while addressing existing challenges.
\vspace{-1em}
\subsubsection{Objective:} Given an input instance $\mathbf{x} \in \mathbb{R}^d$ and a trained ML model \linebreak $f: \mathbb{R}^d \rightarrow \mathbb{R}$, the task is to generate a set of $k$ CF examples ${\mathbf{c}_1, \mathbf{c}_2, \dots, \mathbf{c}_k}$ that influence the model’s decision toward a specified target behavior while maintaining interpretability and actionability.
\vspace{-1em}
\subsubsection{Input Components:}
The CF generation problem is formally defined by:
\begin{itemize}
\item \textbf{$\mathbf{x}$}: The original input instance, represented as a $d$-dimensional feature vector.
\item \textbf{$f$}: A trained ML model, assumed to be differentiable, mapping inputs to real-valued predictions.
\item \textbf{$k$}: The number of CF examples to generate.
\item \textbf{$\mathcal{C}$}: The feasible space of CF candidates, restricted by domain constraints.
\end{itemize}
\vspace{-1em}
\subsubsection{Output:}
A set of CF examples ${\mathbf{c}_1, \mathbf{c}_2, \dots, \mathbf{c}_k} \subseteq \mathcal{C}$, each adhering to defined optimization criteria.
\vspace{-1em}
\subsection{Hypotheses}
The generation of CF explanations follows the hypotheses:
\begin{enumerate}
\item \textbf{Feasibility Hypothesis}: CF explanations close to the original input $\mathbf{x}$ in feature space are more actionable for users.
\item \textbf{Diversity Hypothesis}: A diverse set of CF explanations improves interpretability.
\item \textbf{Robustness Hypothesis}: CFs that remain stable under small perturbations to $\mathbf{x}$ are critical for ensuring reliability.
\end{enumerate}
\vspace{-1em}
\subsection{Mathematical Formalizations}
To meet these hypotheses, the generation process is formulated as an optimization problem. Let $\mathbf{c}_i \in \mathcal{C}$ denote a CF candidate.  The objective is to minimize the following proposed loss function:
\small
\[
\mathcal{L}(\mathbf{c}) =
\operatorname*{arg\,min}_{c_1,\ldots,c_k}
\bigl(
  \mathcal{L}_{y\_{\text{loss}}}
  + \lambda_p \mathcal{L}_{\text{Proximity}}
  - \lambda_d \mathcal{L}_{\text{Diversity}}
  - \lambda_r \mathcal{L}_{\text{Robustness}}
\bigr),
\]
\normalsize

\noindent where $\lambda_p$, $\lambda_d$, and $\lambda_r$ are hyperparameters that balance proximity, diversity, and robustness, respectively.

\subsubsection{Proximity Metric:}
The proximity metric enables minimal deviation from $\mathbf{x}$, computed by $\frac{1}{k} \sum_{i=1}^k | \mathbf{c}_i - \mathbf{x} |_p$ which is a norm-based distance metric. Proximity can be incorporated into the objective function as a penalty term to minimize unnecessary deviations from the original input with $\mathcal{L}_{\text{Proximity}} = \lambda_p \cdot \| \mathbf{c}_i - \mathbf{x} \|_p,$ where $\lambda_p$ is a hyperparameter that controls the trade-off between proximity and other objectives. Here, $\lambda_p$ is set to 0.5, matching the value used in the original DiCE framework to enable a fair comparison with the benchmark method DiCE \cite{Mothilal2020}.

\subsubsection{Sparsity Metric:} The sparsity metric quantifies the number of features that are needed to alter the output of a black-box model. In this study, the number of different features between the input instance and the generated CF is measured using this metric \cite{Mothilal2020}. The sparsity is defined as the average of the number of different features deducted from 1 which can be written mathematically by $1 - \frac{1}{k \times d}\,
  \sum_{i=1}^{k}\sum_{j=1}^{d}
  \mathbf{1}\!\left[c_{i}^{\,j} \neq x^{\,j}\right]$.

\subsubsection{Diversity Metric:}
Diversity among CF explanations is defined with $\det(\mathbf{K})$ where $\mathbf{K}$ refers to a similarity kernel matrix. The kernel matrix is defined by $\mathbf{K}_{i,j} = 1/ [{1 + \text{dist} \,(\mathbf{c}_i, \mathbf{c}_j)]}$, where $\text{dist}(\cdot, \cdot)$ is a distance metric such as $\ell_1$ or $\ell_2$ norm to measure the difference between CF explanations $\mathbf{c}_i$ and $\mathbf{c}_j$.

Diversity can be explicitly incorporated into the loss function to encourage variability among CF explanations which can be formulated by \linebreak $\mathcal{L}_{\text{Diversity}} = -\lambda_d \cdot \det(\mathbf{K})$, where $\lambda_d$ is a hyperparameter that regulates the trade-off between diversity and other objectives. The higher the determinant value is, there are more diverse the set of CF explanations. Following the DiCE framework \cite{Mothilal2020}, $\lambda_d = 1.0$ is set to 1.0.

\subsubsection{Robustness Metric:}
To compute the robustness metric, we adopt the \linebreak Dice–Sørensen coefficient, defined as $\displaystyle\frac{2 \cdot |\mathbf{c} \cap \mathbf{c}'|}{|\mathbf{c}| + |\mathbf{c}'|}$. In this context, $\mathbf{c}$ denotes the CF explanations generated for the input $\mathbf{x}$, while $\mathbf{c}'$ represents the perturbed CF explanations, given by $\mathbf{c}' = \mathbf{c} + \boldsymbol{\delta}$, where $\boldsymbol{\delta}$ denotes the perturbation applied to the generated CFs.

During the computation of the robustness loss, both the CFs and their perturbed counterparts are binarized to ensure discrete representations. The Dice--Sørensen coefficient is then used to measure the distance between the resulting binary vectors. Robustness is incorporated into the overall objective function through a regularization term defined as:
\[
\mathcal{L}_{\text{Robustness}} = \lambda_r \cdot \text{robustness\_loss},
\]
where $\lambda_r$ is a tunable hyperparameter and \texttt{robustness\_loss} corresponds to the Dice--Sørensen distance between the original and perturbed CF explanations.

To determine an appropriate value for $\lambda_r$, we conducted a grid search within the range $[0, 1]$ across all datasets. For example, on the \textit{Adult Income} dataset using the PyTorch backend, the robustness loss was found to peak at \linebreak $\lambda_r = 0.4$, although the total loss continued to decrease beyond this point. Moreover, the variation in the computed robustness across the grid was relatively small, approximately $3 \times 10^{-4}$. Based on these findings, we selected $\lambda_r = 0.4$ as a moderate and balanced value, ensuring that robustness contributes meaningfully to the overall optimization without dominating the objectives of proximity and diversity. The same grid search procedure was applied to all datasets.


\subsection{Dataset Information}
To assess the robustness of our CF explanation framework, we conduct evaluations across four diverse benchmark datasets.

\begin{itemize}
\item \textbf{Adult-Income} \cite{adult_2}: Extracted from the UCI Machine Learning Repository, this dataset includes demographic and employment-related features. We use 8 key attributes, such as hours per week, education level, and occupation, to classify individuals earning above or below \$50,000 per year.
\item \textbf{LendingClub} \cite{openintro_lending_club}: This dataset contains five years of loan records (2007-2011) from LendingClub. We select 8 essential features, including employment years, annual income, and credit history, to predict whether a borrower will repay their loan.
\item \textbf{German-Credit} \cite{statlog_(german_credit_data)_144}: This dataset includes 20 financial and demographic attributes related to loan applicants. The task is to classify applicants as good or bad credit risks.
\item \textbf{COMPAS} \cite{openml_compas_45039}: A dataset from ProPublica focused on recidivism prediction in the U.S. criminal justice system. After preprocessing, 5 key features are retained (\textit{e.g.}, age, prior offenses, and charge severity) to predict whether a defendant will reoffend within two years.
\end{itemize}

\subsection{Experimental Setup for Robustness Assessment}
To benchmark \textbf{DiCE-Extended} against the original \textbf{DiCE} explainer, we implemented a unified evaluation pipeline across four tabular datasets (Adult-Income, Lending-Club, German-Credit, and COMPAS-Recidivism) and three model back-ends: \texttt{Scikit-learn}, \texttt{PyTorch}, and \texttt{TensorFlow}.

\vspace{-1em}
\subsubsection{Model Architectures}

The experimental setup incorporates two distinct model architectures, each tailored to evaluate CF explanations from different learning paradigms:

\vspace{-0.5em}
\paragraph{Tree-based Model.} We utilize a Random Forest classifier implemented with the \texttt{Scikit-learn} library. The model operates as a non-parametric ensemble learning method, aggregating multiple decision trees to enhance predictive stability. The hyperparameter settings enable reproducibility across experiments, with 100 estimators, the Gini criterion, and no maximum depth. 

\paragraph{Neural Network Model.} To explore CF explanations in DL environments, we implement a two-layer feedforward neural network in both \texttt{PYT} and \texttt{TF2}. The architecture consists of a fully connected hidden layer with ReLU activation, followed by an output layer utilizing a Sigmoid activation function:
\begin{equation}
f(\mathbf{x}) = \sigma ( W_2 \cdot \text{ReLU}(W_1 \mathbf{x} + b_1) + b_2 ),
\end{equation}
where $W_1 \in \mathbb{R}^{n \times d}$ and $W_2 \in \mathbb{R}^{1 \times n}$ are the weight matrices. Here $n$ is the number of features, $b_1 \in \mathbb{R}^{n}$ and $b_2 \in \mathbb{R}$ are bias terms, and $\sigma$ represents the Sigmoid activation function. This neural network is trained via backpropagation using a binary cross-entropy loss function. The chosen architecture enables model flexibility while maintaining computational efficiency during CF generation.

\vspace{-1em}
\subsubsection{Training Configuration}
For consistent model evaluation, we follow a standardized training protocol across all back-end frameworks. The training process employs the following hyperparameter configurations:

\begin{itemize}
\item \textbf{Learning Rate:} Set to 0.001 to promote stable gradient updates and mitigate convergence issues.
\item \textbf{Epochs:} Fixed at 10 iterations, balancing training duration and model generalization.
\item \textbf{Data Partitioning:} An 80\%-20\% train-test split is applied, maintaining class distribution integrity within each dataset.
\item \textbf{Batch Processing:} Mini-batch gradient descent is utilized, with a batch size of 16 for training and 4 for testing.
\item \textbf{Optimization Strategy:} The Adam optimizer is employed, leveraging adaptive moment estimation to dynamically adjust learning rates.
\end{itemize}

To prevent overfitting, early stopping mechanisms are employed by monitoring the validation loss and terminating training when no further improvement is observed. Additionally, evaluation metrics such as accuracy and loss progression are continuously logged to track model performance and assess the stability of CF explanations.

\section{Experiments and Results}
\label{sec:4-evaluation}
\subsection{Loss Progression in CF Explainers}
\begin{figure}[h!]
\centering
\includegraphics[width=\textwidth]{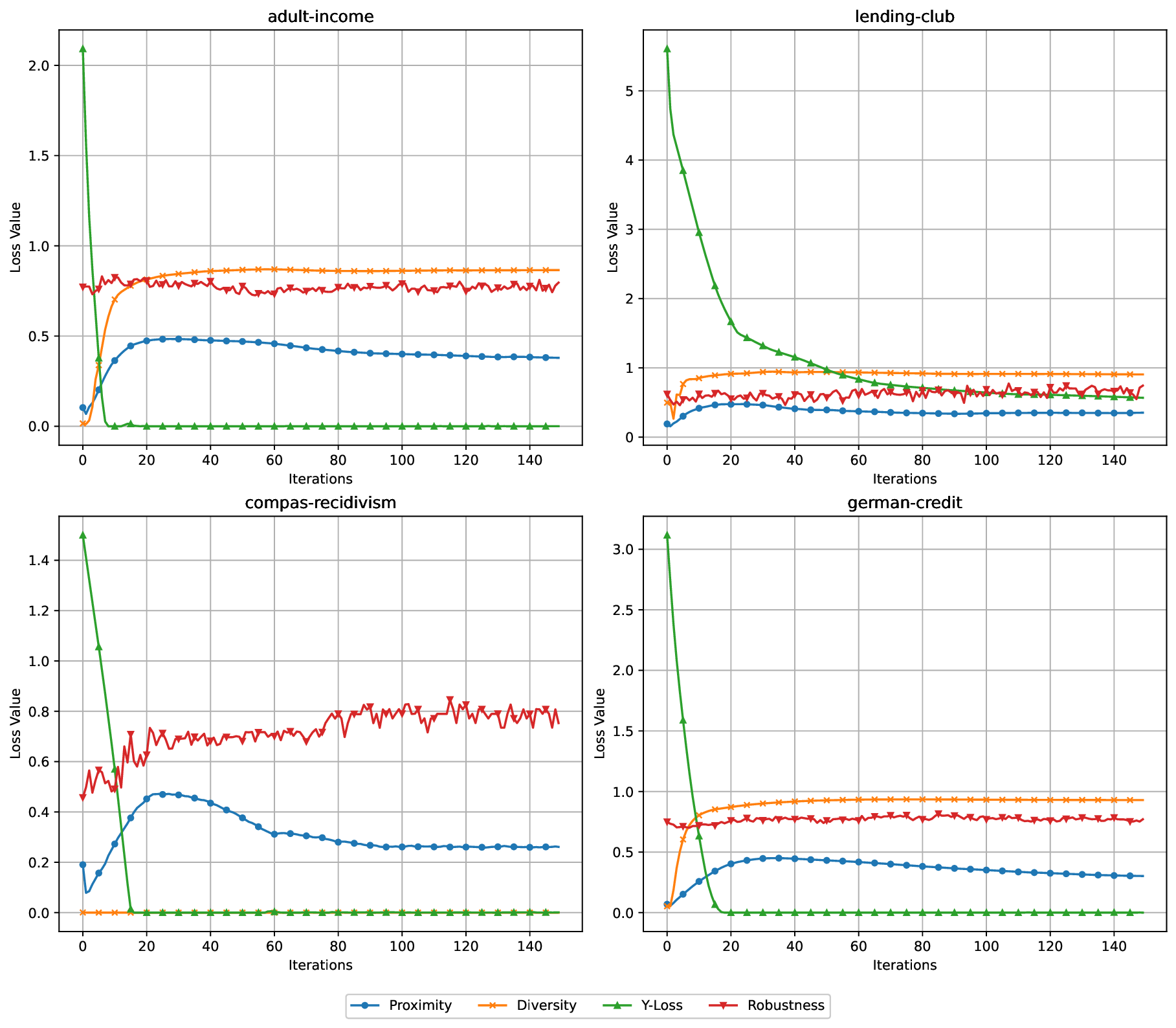}
\caption{Loss trends in DiCE-Extended across datasets for the TensorFlow backend, illustrating the optimization of class loss, proximity loss, diversity loss, and robustness loss during CF generation.}
\label{fig:explainer_loss_tf2_dice_x}
\end{figure}

We analyzed the evolution of various loss components, including robustness loss, during the CF generation process across four datasets. Figure~\ref{fig:explainer_loss_tf2_dice_x} illustrates how the optimization procedure reduces these losses over successive iterations, leading to stable and interpretable CF explanations. Focusing on the robustness loss, we observe initial fluctuations during the early iterations, followed by a gradual and steady decline as the optimization progresses. By approximately the 50\textsuperscript{th} iteration, the robustness loss stabilizes, indicating that the generated CFs become increasingly consistent under small perturbations of input features. This behavior demonstrates that the optimization framework effectively enhances the resilience of CF explanations, improving their reliability in real-world applications.

The convergence of robustness loss is aligned with the overall minimization strategy, which simultaneously balances proximity, diversity, and stability without compromising interpretability. Furthermore, the observed stabilization suggests that after a certain number of iterations, further optimization yields diminishing returns in terms of robustness improvement. This finding highlights the efficiency of DiCE-Extended in generating robust CFs within a limited number of optimization steps, making it suitable for time-sensitive and decision-critical applications.

\subsection{DiCE-Extended vs. DiCE: Counterfactual Quality and Fidelity}

To assess the added value of DiCE-Extended, we evaluated several performance metrics, grouped into the following categories:

\textbf{1. Counterfactual Quality Suite.} For each method, we report standard evaluation metrics, including \emph{validity}, \emph{proximity}, \emph{sparsity}, \emph{diversity}, and the proposed \emph{robustness} measure, as illustrated in Figure~\ref{fig:explainer_loss_tf2_dice_x}.

\textbf{2. Local Fidelity.} To evaluate local fidelity, we trained a 1-nearest-neighbour (1-NN) surrogate model on the union of the original query point and its generated CFs. The surrogate’s accuracy was then measured on 1000 synthetic neighbors sampled and scaled using the mean absolute deviation (MAD) around the query. The MAD is defined as:
\[
\text{MAD} = \frac{1}{n}\sum_{i=1}^{n} \lvert x_i - m(X) \rvert,
\]
where $m(X)$ represents the mean of the dataset $X$. Comparative local fidelity results for DiCE-Extended and the benchmark DiCE method are reported in Table~\ref{tab:_1_nn_accuracy_dice_x} and Table~\ref{tab:_1_nn_accuracy_dice}. Higher 1-NN accuracy indicates that the CF set more accurately reflects the model’s local decision boundary~\cite{Mothilal2020}.

\paragraph{Counterfactual Generation.} For each \emph{(dataset, backend)} pair, we randomly sampled $N=50$ test instances and generated $k=10$ CFs for each instance, targeting the opposite class (\emph{desired\_class}~$=1 - y$).

\paragraph{Statistical Analysis.} For each instance, metric differences between DiCE-Extended and DiCE
\[
\bigl(\text{DiCE-Extended} - \text{DiCE}\bigr)
\]
were analyzed using two-tailed \emph{paired} $t$-tests ($H_0$: mean difference $= 0$; $\alpha = 0.05$). $p$-values are reported in scientific notation, and values below 0.05 are highlighted in bold to indicate statistical significance.

\vspace{-1em}

\subsection{Performance Gains Across Counterfactual Dimensions}

\textbf{DiCE-Extended} consistently generated \emph{closer yet valid} CFs, as shown in Table~\ref{tab:_1_nn_accuracy_dice_x}. Across all four datasets and three model back-ends, the median proximity improved significantly, shrinking by 30\%–95\% while maintaining a validity rate of \(100\%\). A two-tailed \textit{paired} \(t\)-test on the 50 per-instance proximity scores confirmed that these reductions were highly significant (\(p < 10^{-5}\)).

\vspace{0.5em}
\noindent
The framework also achieved notable improvements in \emph{diversity} and \emph{sparsity}. On high-dimensional datasets such as \textsc{Adult} and \textsc{Lending-Club}, CF diversity nearly doubled. Concurrently, Sparsity, defined as the average number of features altered, showed improvements of up to \(50\%\), with all differences statistically significant at the \textit{p}-value threshold of \(< 0.01\).

\vspace{0.5em}
\noindent
Improvements were most pronounced in the newly introduced \textbf{robustness} metric. For PyTorch and TensorFlow models, DiCE-Extended demonstrated absolute gains of 0.05–0.13 in robustness scores, all statistically significant at the \(p < 0.01\) level. These results suggest that the additional robustness loss effectively safeguards CF explanations against input perturbations. Although the gains were smaller for tree-based Scikit-learn models, which are inherently more stable, they were directionally consistent.

\vspace{0.5em}
\noindent
Importantly, these enhancements were achieved with only \emph{minimal computational overhead} due to vectorized implementation. On average, generation time increased by approximately 1.2×, yet each configuration successfully produced the required number of valid CFs per query without degradation in performance.

\vspace{0.5em}
\noindent
The \textit{1-NN fidelity evaluation} further demonstrated that DiCE-Extended preserves, and in many cases improves, the alignment of CFs with the model’s local decision boundary. Specifically, 1-NN fidelity increased by 5–11\% for Scikit-learn models on the \textsc{Adult-Income} and \textsc{Lending-Club} datasets, with statistical significance at the 2\% level. For TensorFlow on \textsc{German-Credit}, improvements of 3–4\% were similarly significant. In contrast, PyTorch on \textsc{COMPAS} showed a modest 7\% advantage for the original DiCE method, but this difference was not statistically significant. In all other configurations, differences were below 4\% and statistically indistinguishable from zero.

\vspace{-1em}

\begin{table}[ht]
\centering
\caption{1-NN fidelity scores (mean $\pm$ standard deviation, $N = 50$) for counterfactual sets generated by DiCE-Extended across three MAD radii (0.5, 1.0, and 2.0).}
\vspace{1em}
\begin{tabularx}{\linewidth}{l l *{3}{>{\centering\arraybackslash}X}}
\hline
\textbf{Dataset} & \textbf{Backend} & \textbf{0.5 MAD} & \textbf{1.0 MAD} & \textbf{2.0 MAD} \\
\hline
Adult-Income & sklearn & 0.54 $\pm$ 0.20 & 0.48 $\pm$ 0.18 & 0.43 $\pm$ 0.16 \\
 & PYT & 0.54 $\pm$ 0.16 & 0.53 $\pm$ 0.16 & 0.51 $\pm$ 0.17 \\
 & TF2 & \textbf{0.54} $\pm$ 0.069 & \textbf{0.54} $\pm$ 0.078 & \textbf{0.54} $\pm$ 0.076 \\
\hline
COMPAS-Recidivism & sklearn & 0.53 $\pm$ 0.12 & 0.53 $\pm$ 0.11 & 0.52 $\pm$ 0.087 \\
 & PYT & 0.57 $\pm$ 0.20 & 0.56 $\pm$ 0.19 & 0.55 $\pm$ 0.16 \\
 & TF2 & \textbf{0.89} $\pm$ 0.076 & \textbf{0.89} $\pm$ 0.074 & \textbf{0.88} $\pm$ 0.075 \\
\hline
German-Credit & sklearn & 0.49 $\pm$ 0.20 & 0.42 $\pm$ 0.20 & 0.38 $\pm$ 0.19 \\
 & PYT & 0.48 $\pm$ 0.13 & 0.48 $\pm$ 0.13 & 0.46 $\pm$ 0.13 \\
 & TF2 & \textbf{0.58} $\pm$ 0.052 & \textbf{0.58} $\pm$ 0.047 & \textbf{0.59} $\pm$ 0.046 \\
\hline
Lending-Club & sklearn & 0.33 $\pm$ 0.16 & 0.26 $\pm$ 0.14 & 0.22 $\pm$ 0.11 \\
 & PYT & 0.46 $\pm$ 0.08 & 0.44 $\pm$ 0.081 & 0.42 $\pm$ 0.078 \\
 & TF2 & \textbf{0.49} $\pm$ 0.056 & \textbf{0.47} $\pm$ 0.055 & \textbf{0.43} $\pm$ 0.067 \\
\hline
\end{tabularx}
\vspace{-1em}
\captionsetup{justification=raggedright,singlelinecheck=false}
\caption*{\footnotesize \small{\textbf{Note:} Highest fidelity per dataset at each MAD radius is highlighted in \textbf{bold}.}}
\label{tab:_1_nn_accuracy_dice_x}
\end{table}

\begin{table}[ht]
\centering
\caption{1-NN fidelity scores (mean $\pm$ standard deviation, $N = 50$) for counterfactual sets generated by DiCE across three MAD radii (0.5, 1.0, and 2.0).}
\vspace{1em}

{\small
\begin{tabularx}{\linewidth}{l l *{3}{>{\centering\arraybackslash}X}}
\hline
\textbf{Dataset} & \textbf{Backend} & \textbf{0.5 MAD} & \textbf{1.0 MAD} & \textbf{2.0 MAD} \\
\hline
Adult-Income & sklearn & 0.43 $\pm$ 0.18 & 0.40 $\pm$ 0.20 & 0.38 $\pm$ 0.21 \\
 & PYT & \textbf{0.58} $\pm$ 0.18 & \textbf{0.57} $\pm$ 0.18 & \textbf{0.56} $\pm$ 0.19 \\
 & TF2 & 0.54 $\pm$ 0.085 & 0.53 $\pm$ 0.080 & 0.51 $\pm$ 0.086 \\
\hline
COMPAS-Recidivism & sklearn & 0.62 $\pm$ 0.14 & 0.56 $\pm$ 0.14 & 0.55 $\pm$ 0.15 \\
 & PYT & 0.64 $\pm$ 0.23 & 0.62 $\pm$ 0.21 & 0.61 $\pm$ 0.18 \\
 & TF2 & \textbf{0.92} $\pm$ 0.081 & \textbf{0.91} $\pm$ 0.085 & \textbf{0.88} $\pm$ 0.10 \\
\hline
German-Credit & sklearn & 0.44 $\pm$ 0.20 & 0.38 $\pm$ 0.21 & 0.35 $\pm$ 0.22 \\
 & PYT & 0.45 $\pm$ 0.15 & 0.44 $\pm$ 0.15 & 0.43 $\pm$ 0.15 \\
 & TF2 & \textbf{0.55} $\pm$ 0.068 & \textbf{0.55} $\pm$ 0.062 & \textbf{0.55} $\pm$ 0.051 \\
\hline
Lending-Club & sklearn & 0.24 $\pm$ 0.11 & 0.20 $\pm$ 0.11 & 0.18 $\pm$ 0.11 \\
 & PYT & \textbf{0.52} $\pm$ 0.084 & \textbf{0.51} $\pm$ 0.076 & \textbf{0.48} $\pm$ 0.078 \\
 & TF2 & 0.47 $\pm$ 0.092 & 0.44 $\pm$ 0.100 & 0.40 $\pm$ 0.100 \\
\hline
\end{tabularx}
}
\vspace{-1em}
\captionsetup{justification=raggedright,singlelinecheck=false}
\caption*{\footnotesize \textbf{Note:} Highest fidelity per dataset at each MAD radius is highlighted in \textbf{bold}.}
\label{tab:_1_nn_accuracy_dice}
\end{table}

\section{Conclusion}
\label{sec:6-conclusion}
Our findings demonstrate that, particularly for DL back-ends, \textbf{DiCE-Extended consistently generates CFs that are closer, sparser, and more diverse, while maintaining near-perfect validity and significantly enhancing robustness}. Robustness scores increased by $+0.05$ to $+0.13$ (with $p < 0.05$), and substantial gains in proximity indicate that the generated CFs more faithfully adhere to the model’s local decision boundaries. Despite a modest increase in generation time (approximately $1.2\times$), each configuration successfully produced the desired number of CFs per query. Future research will focus on fine-tuning the trade-offs among proximity, diversity, and robustness; integrating domain-specific constraints; and extending the framework with robust data privacy guarantees, thereby enabling privacy-aware CF generation in real-world applications.

\paragraph{\textbf{Disclosure of Interests.}} The authors have no competing interests to declare that are relevant to the content of this article.

\FloatBarrier
\vspace{5em}
\bibliographystyle{splncs04}
\bibliography{references}

\end{document}